\documentclass[11pt,a4paper]{article}
\usepackage[hyperref]{acl2020}
\usepackage{times}
\usepackage{latexsym}
\usepackage{amsmath}
\usepackage{times}
\usepackage{helvet}
\usepackage{courier}
\usepackage{natbib}
\usepackage{amssymb}
\usepackage{amsfonts}
\usepackage{bbm}
\usepackage{todonotes}
\usepackage{hyperref}
\usepackage{mathtools}
\usepackage{booktabs}
\usepackage{array}
\usepackage{float}
\usepackage{ulem}
\usepackage{algorithm}
\usepackage{algorithmic}

\usepackage{microtype}

\aclfinalcopy 
\def\aclpaperid{3276} 

\newcommand{\real}{\ensuremath{\mathbb{R}}}

\newcommand{\KL}[2]{\ensuremath{\mathcal{KL}({#1} || {#2})}}

\newcolumntype{C}{ >{\centering\arraybackslash} m{0.5cm} }
\newcolumntype{D}{ >{\centering\arraybackslash} m{2.8cm} }
\newcolumntype{E}{ >{\centering\arraybackslash} m{2cm} }

\title{A Relaxed Matching Procedure for Unsupervised BLI}

\author{Xu Zhao$^1$, Zihao Wang$^1$, Hao Wu$^2$, Zhang Yong$^1$ \footnotemark[2] \\
  $^1$BNRist, Department of Computer Science and Technology, RIIT,\\
	Institute of Internet Industry, Tsinghua University, Beijing, China \\
  $^2$Department of Mathematical Sciences, Tsinghua University, Beijing, China\\
  \texttt{\{zhaoxu18, wzh17\}@mails.tsinghua.edu.cn} \\
  \texttt{\{hwu, zhangyong05\}@tsinghua.edu.cn}}

\begin{document}
\maketitle
\renewcommand{\thefootnote}{\fnsymbol{footnote}}
\footnotetext[2]{Yong Zhang is the corresponding author.}
\begin{abstract}
Recently unsupervised \textbf{Bilingual Lexicon Induction(BLI)} without any parallel corpus has attracted much research interest. One of the crucial parts in methods for the BLI task is the matching procedure. Previous works impose a too strong constraint on the matching and lead to many counterintuitive translation pairings. Thus, We propose a relaxed matching procedure to find a more precise matching between two languages. We also find that aligning source and target language embedding space bidirectionally will bring significant improvement. We follow the previous iterative framework to conduct experiments. Results on standard benchmark demonstrate the effectiveness of our proposed method, which substantially outperforms previous unsupervised methods.
\end{abstract}

\section{Introduction}
Pretrained word embeddings \citep{DBLP:conf/nips/MikolovSCCD13} are the basis of many other natural language processing and machine learning systems. Word embeddings of a specific language contain rich syntax and semantic information. \citet{DBLP:journals/corr/MikolovLS13} stated that the continuous embedding spaces exhibit similar structures across different languages, and we can exploit the similarity by a linear transformation from source embedding space to target embedding space. This similarity derives the \textbf{Bilingual Lexicon Induction(BLI)} task. The goal of bilingual lexicon induction is to align two languages' embedding space and generates word translation lexicon automatically. This fundamental problem in natural language processing benefits much other research such as sentence translation \citep{rapp1995identifying, fung1995compiling}, unsupervised machine translation \citep{lample2017unsupervised}, cross-lingual information retrieval \citep{lavrenko2002cross}.

Recent endeavors \cite{DBLP:conf/iclr/LampleCRDJ18, DBLP:conf/emnlp/Alvarez-MelisJ18, DBLP:conf/aistats/GraveJB19, artetxe2017learning} have proven that unsupervised BLI's performance is even on par with the supervised methods. A crucial part of these approaches is the \textbf{matching procedure}, i.e., how to generate the translation plan.  \citet{DBLP:conf/emnlp/Alvarez-MelisJ18} used Gromov-Wasserstein distance to approximate the matching between languages. \citet{DBLP:conf/aistats/GraveJB19} regarded it as a classic optimal transport problem and used the sinkhorn algorithm \citep{DBLP:conf/nips/Cuturi13} to compute the translation plan.

In this work, we follow the previous iterative framework but use a different matching procedure. Previous iterative algorithms required to compute an approximate 1 to 1 matching every step. This 1 to 1 constraint brings out many redundant matchings. Thus in order to avoid this problem, we relax the constraint and control the relaxation degree by adding two KL divergence regularization terms to the original loss function. This relaxation derives a more precise matching and significantly improves performance. Then we propose a bidirectional optimization framework to optimize the mapping from source to target and from target to source simultaneously. In the section of experiments, we verify the effectiveness of our method, and results show our method outperforms many SOTA methods on the BLI task.

\section{Background}
The early works for the BLI task require a parallel lexicon between languages. Given two embedding matrices $X$ and $Y$ with shape $n \times d$ ($n$:word number, $d$:vector dimension) of two languages and word $x_i$ in $X$ is the translation of word $y_i$ in $Y$, i.e., we get a parallel lexicon $X  \rightarrow Y$.  \citet{DBLP:journals/corr/MikolovLS13} pointed out that we could exploit the similarities of monolingual embedding spaces by learning a linear transformation $W^{\star}$ such that
\begin{equation}
W^{\star}= \mathop{\arg\min}_{W \in M_{d}(\mathbb{R})}\|XW-Y\|^{2}_{F}
\end{equation}
where $M_{d}(\mathbb{R})$ is the space of $d \times d$ matrices of real numbers. \citet{DBLP:conf/naacl/XingWLL15} stated that enforcing an orthogonal constraint on $W$ would improve performance. There is a closed-form solution to this problem called \textbf{Procrutes}: $W^{\star}=Q=UV^T  $ where $USV^T=XY^T$. 

Under the unsupervised condition without parallel lexicon, i.e., vectors in $X$ and $Y$ are totally out of order, \citet{DBLP:conf/iclr/LampleCRDJ18} proposed a domain-adversarial approach for learning $W^{\star}$. On account of the ground truth that monolingual embedding spaces of different languages keep similar spatial structures, \citet{DBLP:conf/emnlp/Alvarez-MelisJ18} applied the Gromov-Wasserstein distance based on infrastructure to find the corresponding translation pairings between $X$ and $Y$ and further derived the orthogonal mapping Q. \citet{DBLP:conf/aistats/GraveJB19} formulated the unsupervised BLI task as
\begin{equation}
\setlength\abovedisplayskip{10pt}
\setlength\belowdisplayskip{10pt}
\min_{Q \in \mathcal{O}_d  , P \in \mathcal{P}_n} \|XQ - PY\|^{2}_{F}
\label{loss}
\end{equation}
where $\mathcal{O}_{d}$ is the set of orthogonal matrices and $\mathcal{P}_n$ is is the set of permutation matrices.Given $Q$, estimating $P$ in Problem~\eqref{loss} is equivalent to the minimization of the 2-Wasserstein distance between the two sets of points: $XQ$ and $Y$.
\begin{equation}
        W^2_2(XQ, Y) = \min_{P \in \mathcal{P}_n}  \langle D, P  \rangle
    \label{W}
\end{equation}
where $D_{ij} = \| x_iQ - y_j\|^2_2$ and $\langle D, P\rangle =\sum_{i, j} P_{ij} D_{ij} $ denotes the matrix inner product.  \citet{DBLP:conf/aistats/GraveJB19} proposed a stochastic algorithm to estimate $Q$ and $P$ jointly. Problem~\eqref{W} is the standard optimal transport problem that can be solved by Earth Mover Distance linear program with $ \mathcal{O}(n^3)$ time complexity.  Considering the computational cost, \citet{zhang2017earth} and \citet{DBLP:conf/aistats/GraveJB19} used the Sinkhorn algorithm~ \citep{DBLP:conf/nips/Cuturi13} to estimate $P$ by solving the entropy regularized optimal tranpsort problem~\citep{peyre2019computational}.

We also take Problem~\eqref{loss} as our loss function and our model shares a similar alternative framework with \citet{DBLP:conf/aistats/GraveJB19}. However, we argue that the permutation matrix constraint on $P$ is too strong, which leads to many inaccurate and redundant matchings between $X$ and $Y$, so we relax it by unbalanced optimal transport.

\citet{DBLP:journals/corr/abs-1811-01124} extended the line of BLI to the problem of aligning multiple languages to a common space. \citet{DBLP:conf/naacl/ZhouMWN19} estimated Q by a density matching method called normalizing flow. \citet{DBLP:conf/aaai/ArtetxeLA18} proposed a multi-step framework of linear transformations that generalizes a substantial body of previous work. \citet{DBLP:journals/corr/abs-1912-01706} further investigated the robustness of \citet{DBLP:conf/aaai/ArtetxeLA18}'s model by introducing four new languages that are less similar to English than the ones proposed by the original paper. \citet{DBLP:conf/acl/ArtetxeLA19a} proposed an alternative approach to this problem that builds on the recent work on unsupervised machine translation.

\section{Proposed Method}
In this section, we propose a method for the BLI task. As mentioned in the background, we take Problem~\eqref{loss} as our loss function and use a similar optimization framework in \citet{DBLP:conf/aistats/GraveJB19} to  estimate $P$ and $Q$ alternatively. Our method focuses on the estimation of $P$ and tries to find a more precise matching $P$ between $XQ$ and $Y$. Estimation of $Q$ is by stochastic gradient descent.  We also propose a bidirectional optimization framework in section 3.2.
\subsection{Relaxed Matching Procedure}

Regarding embedding set $X$ and $Y$ as two discrete distributions 
$\mu = \sum_{i=1}^I u_i \delta_{x_i}$ and 
$\nu = \sum_{j=1}^J v_j \delta_{y_j}$, where $u$ (or $v$) is column vector satisfies $\sum_i u_i = 1, u_i > 0$ ($v$ is similar), $\delta_x$ is the Dirac function supported on point $x$.

Standard optimal transport enforces the optimal transport plan to be the joint distribution $P \in \mathcal{P}_n$. This setting leads to the result that every mass in $\mu$ should be matched to the same mass in $\nu$. Recent application of unbalanced optimal transport ~\citep{DBLP:journals/corr/abs-1904-10294} shows that the relaxation of the marginal condition could lead to more flexible and local matching, which avoids some counterintuitive matchings of source-target mass pairs with high transportation cost.

The formulation of unbalanced optimal transport~\citep{chizat2018interpolating} differs from the balanced optimal transport in two ways. Firstly, the set of transport plans to be optimized is generalized to $\real_+^{I\times J}$. Secondly, the marginal conditions of the Problem~\eqref{W} are relaxed by two KL-divergence terms.
\begin{equation}
    \setlength \abovedisplayskip{0em}
    \begin{split}
          \min_{P \in \real_+^{I\times J}}  \langle D, P\rangle 
     + \lambda_1 & \KL{P \mathbbm{1}_J}{u} \\
    & + \lambda_2 \KL{P^T \mathbbm{1}_I}{v}
    \end{split}
    \label{RelaxedW}
\end{equation}
where $\KL{p}{q} = \sum_i p_i\log\left(\frac{p_i}{q_i}\right) - p_i + q_i$ is the KL divergence.

We estimate $P$ by considering the relaxed Problem~\eqref{RelaxedW} instead of the original Problem~\eqref{W} in~\citep{DBLP:conf/aistats/GraveJB19}.
Problem~\eqref{RelaxedW} could also be solved by entropy regularization with the generalized Sinkhorn algorithm~\citep{chizat2018scaling,DBLP:journals/corr/abs-1904-10294,peyre2019computational}.
\begin{algorithm}[t]
\caption{Generalized Sinkhorn Algorithm}
\begin{algorithmic}[1]
    \REQUIRE source and target measure $\mu_i\in \real^m_+, \nu_j\in \real^n$, entropy regularizer $\epsilon$, KL relaxation coefficient $\lambda_1,\lambda_2$ and distance matrix$D_{ij}$.
    \ENSURE Transport Plan$P_{ij}$
    \STATE Initialize $u \gets 0 \in \real^m$, $v \gets 0\in \real^n$, $K \gets e^{-D/\gamma} \in \real^{m\times n}$
    \WHILE{not converge}
    \STATE $u \gets \left(\frac{\mu}{Kv}\right)^{\frac{\lambda_1}{\epsilon+\lambda_1}}$
    \STATE $v \gets \left(\frac{\nu}{K^\top u}\right)^{\frac{\lambda_2}{\epsilon+\lambda_2}}$
    \ENDWHILE
    \STATE $P \gets \text{diag}(u) K \text{diag}(v)$
\end{algorithmic}
\end{algorithm}

In short, we already have an algorithm to obtain the minimum of the Problem~\eqref{RelaxedW}. In order to avoid the hubness phenomenon, we replace $\mathcal{l}_2$ distance of embedding with the $rcsls$ distance proposed in \citet{DBLP:conf/emnlp/JoulinBMJG18} formalized as $D_{ij} = rcsls(x_iQ, y_j)$.
$rcsls$ can not provide significantly better results than euclidean distance in our evaluation. However, previous study suggests that RCSLS could be considered as a better metric between words than euclidean distance. So we propose our approach with RCSLS. The "relaxed matching" procedure and  the "bi-directional optimization" we proposed bring most of the improvement.

We call this relaxed estimation of $P$ as \textbf{Relaxed Matching Procedure(RMP)}. With RMP only when two points are less than some radius apart from each other, they may be matched together. Thus we can avoid some counterintuitive matchings and obtain a more precise matching $P$. In the section of experiments we will verify the effectiveness of RMP.

\subsection{Bidirectional Optimization}
Previous research solved the mapping $X$ to $Y$ and the mapping $Y$ to $X$ as two independent problems, i.e., they tried to learn two orthogonal matrix $Q_1$ and $Q_2$ to match the $XQ_1$ with $Y$ and $YQ_2$ with $X$, respectively. Intuitively from the aspect of point cloud matching, we consider these two problems in opposite directions are symmetric. Thus we propose an optimization framework to solve only one $Q$ for both directions. 

In our approach, we match $XQ$ with $Y$ and $YQ^T$ with $X$ simultaneously. Based on the stochastic optimization framework of \citet{DBLP:conf/aistats/GraveJB19}, we randomly choose one direction to optimize at each iteration.

The entire process of our method is summarized in Algorithm~\ref{alg:1}. At iteration $i$, we start with sampling batches $X_b$, $Y_b$ with shape $\mathbb{R}^{b \times d}$. Then we generate a random integer $rand$ and choose to map $X_bQ$ to $Y_b$ or map $Y_bQ^T$ to $X_b$ by $rand$'s parity. Given the mapping direction, we run the RMP procedure to solve Problem~\eqref{RelaxedW} by sinkhorn and obtain a matching matrix $P^*$ between $X_bQ$ and $Y_b$(or $Y_bQ^T$ and $X$). Finally we use gradient descent and procrutes to update $Q$ by the given $P^*$. The procedure of $Q$'s update is detailed in \citet{DBLP:conf/aistats/GraveJB19}.

\begin{algorithm}[t]
\caption{Bidirectional Optimization with RMP}
\label{alg:1}
\begin{algorithmic}[1]
\REQUIRE word vectors from two languages$X$, $Y$
\ENSURE Transformation $Q$
\FOR{each $e \in [1,E]$}
\FOR{each $i \in [1,I]$}
\STATE Draw $X_b$, $Y_b$ of size $b$ from $X$ and $Y$
\STATE set $rand = random()$
\IF{$rand \mod 2 = 1$} 
\STATE $Y_b, X_b, Q \Leftarrow X_b, Y_b, Q^T$
\ENDIF 
\STATE Run RMP by solving Problem~\eqref{RelaxedW} and  obtain $P^*$\
\STATE Update Q by gradient descent and Procrutes
\IF{$rand \mod 2 = 1$} 
\STATE $Q \Leftarrow Q^T$
\ENDIF
\ENDFOR
\ENDFOR
\end{algorithmic}

\end{algorithm}
\vspace{-0.3cm}

\section{Experiments}

In this section, we evaluate our method in two settings. First, We conduct distillation experiments to verify the effectiveness of RMP and bidirectinal optimization.
Then we compare our method consisting of both RMP and bi-directional optimization with various SOTA methods on the BLI task.

\begin{table*}[]
	\begin{center}
	\renewcommand{\arraystretch}{1.5}
	\begin{tabular}{DECCCCCCCCCCCC}
		\toprule[2pt]
		Method &             & \multicolumn{2}{c}{EN-ES} & \multicolumn{2}{c}{EN-FR} & \multicolumn{2}{c}{EN-DE} & \multicolumn{2}{c}{EN-RU} &  \multicolumn{2}{c}{EN-IT} & Avg.\\ \hline
		& Supervision &    $\rightarrow$    &    $\leftarrow$         &       $\rightarrow$      &      $\leftarrow$       &      $\rightarrow$       &      $\leftarrow$             &     $\rightarrow$        &    $\leftarrow$         &      $\rightarrow$       &      $\leftarrow$       \\ \hline
		Proc.   & 5K words    & 81.9        & 83.4        & 82.1        & 82.4        & 74.2        & 72.7        & 51.7        & 63.7    & 77.4  & 77.9 &74.7      \\
		RCSLS & 5K words & 84.1 &86.3 &83.3 &84.1 &79.1 &76.3 &57.9 &67.2 &&& 77.3 \\ \hline
		GW & None        & 81.7        & 80.4        & 81.3        & 78.9        & 71.9        & \textbf{78.2}        & 45.1        & 43.7         &  78.9       & 75.2  & 71.5\\ 
		Adv. -  Refine & None        & 81.7        & 83.3        & 82.3        & 82.1        & 74.0        & 72.2        & 44.0        & 59.1        &  77.9       & 77.5  & 73.4\\ 
		W.Proc. - Refine & None        & \textbf{82.8}        & 84.1        & 82.6        & 82.9        & 75.4        & 73.3        & 43.7        & 59.1          &         &  &73.0\\  
		Dema - Refine & None & 82.8 & 84.9 & 82.6 & 82.4 & 75.3 & 74.9 & 46.9 & 62.4 &&& 74.0\\ \hline
		Ours - Refine & None &  82.7 & \textbf{85.8 } &
		\textbf{83.0 } & \textbf{83.8} & \textbf{76.2} & 74.9 & \textbf{48.1} & \textbf{64.7} & \textbf{79.1} & \textbf{80.3} & \textbf{75.9}\\ 
		\bottomrule[2pt]     
		
	\end{tabular}
	\end{center}
	\caption{Comparison between SOTA methods on BLI task. Methods in Line 1-2 are supervised. Methods in Line 3-8 are unsupervised. Except the GW’ method, other unsupervised methods are refined. In bold, the best among unsupervised approaches. All numbers of others are taken from
their papers. ('EN': English, 'ES': Spanish, 'FR': French, 'DE': German, 'RU': Russian, 'IT': Italian).}
	\label{Table1}
\end{table*}

\textbf{DataSets\footnote{https://github.com/facebookresearch/MUSE}}  We conduct word translation experiments on 6 pairs of languages and use pretrained word embedding from fasttext. We use the bilingual dictionaries opensourced in the work \cite{DBLP:conf/iclr/LampleCRDJ18} as our evaluate set.We use the CSLS retrieval method for evaluation as \citet{DBLP:conf/iclr/LampleCRDJ18} in both settings. All the translation accuracy reported is the precision at 1 with CSLS criterion. We open the source code on Github\footnote{https://github.com/BestActionNow/bidirectional-RMP}.

\subsection{Main Results}

Through the experimental evaluation, we seek to demonstrate the effectiveness of our method compared to other SOTA methods. The word embeddings are normalized and centered before entering the model. We start with a batch size 500 and 2000 iterations each epoch. We double the batch size and quarter the iteration number after each epoch. First 2.5K words are taken for initialization, and samples are only drawn from the first 20K words in the frequently ranking vocabulary. The coefficients $\lambda_1$ and $\lambda_2$ of the relaxed terms in Problem~\eqref{RelaxedW} are both set to 0.001.

\textbf{Baselines} We take basic Procrutes and RCSLS-Loss of \citet{DBLP:conf/emnlp/JoulinBMJG18} as two supervised baselines. 
Five unsupervised methods are also taken into accounts: 
the Gromov Wasserstein matching method of \citet{DBLP:conf/emnlp/Alvarez-MelisJ18},
the adversarial training(Adv.-Refine) of \citet{DBLP:conf/iclr/LampleCRDJ18},  the Wasserstein Procrutes method(W.Proc.-Refine) of  \citet{DBLP:conf/aistats/GraveJB19}, the density matching method(Dema-Refine) of \citet{DBLP:conf/naacl/ZhouMWN19}.

In Table~\ref{Table1}, it's shown that leading by an average of 2 percentage points, our approach outperforms other unsupervised methods in most instances and is on par with the supervised method on some language pairs. Surprisingly we find that our method achieves significant progress in some tough cases such as English - Russian, English - Italian, which contain lots of noise. Our method guarantees the precision of mapping computed every step which achieves the effect of noise reduction. 

However, there still exists an noticeable gap between our method and the supervised RCSLS method, which indicates  further research can be conducted to absorb the superiority of this metric to unsupervised methods.

We also compare our method with W.Proc on two non-English pairs including FR-DE and FR-ES to show how bidirectional relaxed matching improves the performance and results are presented in Table~\ref{Table2}. Most of the recent researches didn't report results of non-English pairs, which makes it hard for fair comparison. However from the results in Table~\ref{Table2}, we could find that our method keeps an advantage over W.Proc. Note that the W.Proc. results here are our implementation rather than that are reported in the original paper.

\begin{table}[t]
    \begin{center}
    \renewcommand{\arraystretch}{1.5}
    \renewcommand{\tabcolsep}{3.0pt}
    \begin{tabular}{lcccc}
    \toprule
          &  FR-DE & DE-FR & FR-ES & ES-FR\\ \hline
         W.Proc.& 65.8 & 73.5 & 82.0 & 84.9\\
         Ours-Refine & 67.7 & 74.0 & 83.3 & 84.9\\
    \bottomrule         
    \end{tabular}   
    \end{center}
    \caption{Comparision bewtween W.Proc. and our method on non-English language pairs}
    \label{Table2}
    \vspace{-0.7cm}
\end{table}

\subsection{Ablation Study}

\begin{figure}
    \centering
    \includegraphics[width=\linewidth]{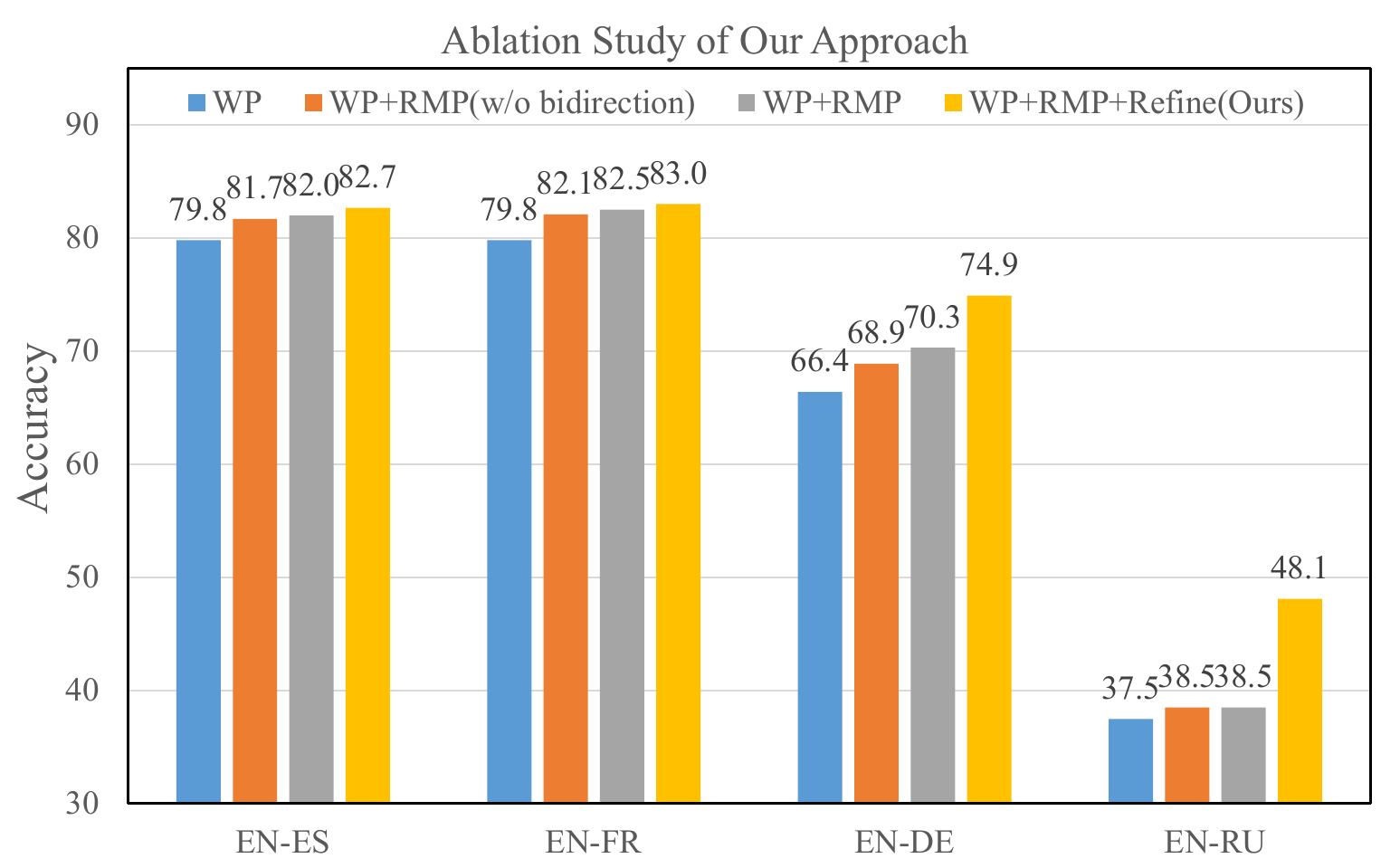}
    \caption{Ablation study of our methods' effectiveness. 'WP' refers to the original Wasserstein Procrutes Method proposed by \citet{DBLP:conf/aistats/GraveJB19}. 'WP-RMP' applies RMP to 'WP'. 'WP-RMP-bidiretion' applies bidirectional optimization framework to 'WP-RMP'. 'WP-RMP-bidirection-refine' applies the refinement procedure to 'WP-RMP-bidirection'.('EN': English, 'ES': Spanish, 'FR': French, 'DE': German, 'RU': Russian, 'IT': Italian).}
    \label{fig:fig1}
    \vspace{-0.3cm}
\end{figure}

The algorithms for BLI could be roughly divided into three parts: 1. initialization, 2 iterative optimization, and 3. refinement procedure, such as \citet{lample2017unsupervised}. W.Proc.\citep{DBLP:conf/aistats/GraveJB19} only covers the first two parts. Our approaches, i.e. relaxed matching and bi-directional optimization are categorized into the second part. To ensure a fair comparison, W.Proc.-Refine is compared to ours-Refine which is discussed in next section. To verify the effectiveness of RMP and bidirectional optimization directly, we apply them to the method proposed in \citet{DBLP:conf/aistats/GraveJB19} one by one. We take the same implementation and hyperparameters reported in their paper and code
\footnote{https://github.com/facebookresearch/fastText/alignment} but using RMP to solve $P$ instead of ordinary 2-Wasserstein.

On four language pairs, We applied RMP, bidirectional optimization and refinement procedure to original W.Proc. gradually and evaluate the performance change. In Figure \ref{fig:fig1} it's clearly shown that after applying bidirectional RMP, the translation accuracy improves by 3 percentage averagely. The results of 'WP-RMP' are worse than 'WP-RMP-bidirection' but better than original 'WP'. Moreover, we find that by applying RMP, a more precise $P$ not only eliminates many unnecessary matchings but also leads to a faster converge of the optimization procedure. Furthurmore, the effectiveness of refinement procedure is quite significant.

To summarize, we consider the average of scores (from en-es to ru-en). By  mitigating the counter-intuitive pairs by polysemies and obscure words, the "relaxed matching" procedure improves the average score about 2 points, the "bi-directional optimization" improves the average score about 0.6 points. From the results we could get some inspiration that our ideas of relaxed matching and bidirectional optimization can also be applied to other frameworks such as adversarial training by \citet{lample2017unsupervised} and Gromov-Wasserstein by \citet{DBLP:conf/emnlp/Alvarez-MelisJ18}.

\section{Conclusion}
This paper focuses on the matching procedure of BLI task. Our key insight is that the relaxed matching mitigates the counter-intuitive pairs by polysemy and obscure words, which is supported by comparing W.Proc.-RMP with W.Proc in Table~\ref{fig:fig1}. The optimal transport constraint considered by W.Proc. is not proper for BLI tasks. Moreover, Our approach also optimizes the translation mapping Q in a bi-directional way, and has been shown better than all other unsupervised SOTA models with the refinement in Table 1.
 
\section{Acknowledgement}
This work was supported by the National Natural Science Foundation of China (11871297, 91646202), National Key R\&D Program of China(2018YFB1404401, 2018YFB1402701), Tsinghua University Initiative Scientific Research Program.

\bibliography{acl2020}

\begin{thebibliography}{23}
\expandafter\ifx\csname natexlab\endcsname\relax\def\natexlab#1{#1}\fi

\bibitem[{Alaux et~al.(2019)Alaux, Grave, Cuturi, and
  Joulin}]{DBLP:journals/corr/abs-1811-01124}
Jean Alaux, Edouard Grave, Marco Cuturi, and Armand Joulin. 2019.
\newblock \href {http://arxiv.org/abs/1811.01124} {Unsupervised hyperalignment
  for multilingual word embeddings}.
\newblock \emph{CoRR}, abs/1811.01124.

\bibitem[{Alvarez{-}Melis and
  Jaakkola(2018)}]{DBLP:conf/emnlp/Alvarez-MelisJ18}
David Alvarez{-}Melis and Tommi~S. Jaakkola. 2018.
\newblock \href {https://aclanthology.info/papers/D18-1214/d18-1214}
  {Gromov-wasserstein alignment of word embedding spaces}.
\newblock In \emph{Proceedings of the 2018 Conference on Empirical Methods in
  Natural Language Processing, Brussels, Belgium, October 31 - November 4,
  2018}, pages 1881--1890.

\bibitem[{Artetxe et~al.(2017)Artetxe, Labaka, and
  Agirre}]{artetxe2017learning}
Mikel Artetxe, Gorka Labaka, and Eneko Agirre. 2017.
\newblock Learning bilingual word embeddings with (almost) no bilingual data.
\newblock In \emph{Proceedings of the 55th Annual Meeting of the Association
  for Computational Linguistics (Volume 1: Long Papers)}, pages 451--462.

\bibitem[{Artetxe et~al.(2018)Artetxe, Labaka, and
  Agirre}]{DBLP:conf/aaai/ArtetxeLA18}
Mikel Artetxe, Gorka Labaka, and Eneko Agirre. 2018.
\newblock Generalizing and improving bilingual word embedding mappings with a
  multi-step framework of linear transformations.
\newblock In \emph{Thirty-Second AAAI Conference on Artificial Intelligence}.

\bibitem[{Artetxe et~al.(2019)Artetxe, Labaka, and
  Agirre}]{DBLP:conf/acl/ArtetxeLA19a}
Mikel Artetxe, Gorka Labaka, and Eneko Agirre. 2019.
\newblock \href {https://doi.org/10.18653/v1/p19-1494} {Bilingual lexicon
  induction through unsupervised machine translation}.
\newblock In \emph{Proceedings of the 57th Conference of the Association for
  Computational Linguistics, {ACL} 2019, Florence, Italy, July 28- August 2,
  2019, Volume 1: Long Papers}, pages 5002--5007.

\bibitem[{Chizat et~al.(2018{\natexlab{a}})Chizat, Peyr{\'e}, Schmitzer, and
  Vialard}]{chizat2018interpolating}
Lenaic Chizat, Gabriel Peyr{\'e}, Bernhard Schmitzer, and Fran{\c{c}}ois-Xavier
  Vialard. 2018{\natexlab{a}}.
\newblock An interpolating distance between optimal transport and fisher--rao
  metrics.
\newblock \emph{Foundations of Computational Mathematics}, 18(1):1--44.

\bibitem[{Chizat et~al.(2018{\natexlab{b}})Chizat, Peyr{\'e}, Schmitzer, and
  Vialard}]{chizat2018scaling}
Lenaic Chizat, Gabriel Peyr{\'e}, Bernhard Schmitzer, and Fran{\c{c}}ois-Xavier
  Vialard. 2018{\natexlab{b}}.
\newblock Scaling algorithms for unbalanced optimal transport problems.
\newblock \emph{Mathematics of Computation}, 87(314):2563--2609.

\bibitem[{Cuturi(2013)}]{DBLP:conf/nips/Cuturi13}
Marco Cuturi. 2013.
\newblock \href
  {http://papers.nips.cc/paper/4927-sinkhorn-distances-lightspeed-computation-of-optimal-transport}
  {Sinkhorn distances: Lightspeed computation of optimal transport}.
\newblock In \emph{Advances in Neural Information Processing Systems 26: 27th
  Annual Conference on Neural Information Processing Systems 2013. Proceedings
  of a meeting held December 5-8, 2013, Lake Tahoe, Nevada, United States},
  pages 2292--2300.

\bibitem[{Fung(1995)}]{fung1995compiling}
Pascale Fung. 1995.
\newblock Compiling bilingual lexicon entries from a non-parallel
  english-chinese corpus.
\newblock In \emph{Third Workshop on Very Large Corpora}.

\bibitem[{Garneau et~al.(2019)Garneau, Godbout, Beauchemin, Durand, and
  Lamontagne}]{DBLP:journals/corr/abs-1912-01706}
Nicolas Garneau, Mathieu Godbout, David Beauchemin, Audrey Durand, and Luc
  Lamontagne. 2019.
\newblock \href {http://arxiv.org/abs/1912.01706} {A robust self-learning
  method for fully unsupervised cross-lingual mappings of word embeddings:
  Making the method robustly reproducible as well}.
\newblock \emph{CoRR}, abs/1912.01706.

\bibitem[{Grave et~al.(2019)Grave, Joulin, and
  Berthet}]{DBLP:conf/aistats/GraveJB19}
Edouard Grave, Armand Joulin, and Quentin Berthet. 2019.
\newblock \href {http://proceedings.mlr.press/v89/grave19a.html} {Unsupervised
  alignment of embeddings with wasserstein procrustes}.
\newblock In \emph{The 22nd International Conference on Artificial Intelligence
  and Statistics, {AISTATS} 2019, 16-18 April 2019, Naha, Okinawa, Japan},
  pages 1880--1890.

\bibitem[{Joulin et~al.(2018)Joulin, Bojanowski, Mikolov, J{\'{e}}gou, and
  Grave}]{DBLP:conf/emnlp/JoulinBMJG18}
Armand Joulin, Piotr Bojanowski, Tomas Mikolov, Herv{\'{e}} J{\'{e}}gou, and
  Edouard Grave. 2018.
\newblock \href {https://www.aclweb.org/anthology/D18-1330/} {Loss in
  translation: Learning bilingual word mapping with a retrieval criterion}.
\newblock In \emph{Proceedings of the 2018 Conference on Empirical Methods in
  Natural Language Processing, Brussels, Belgium, October 31 - November 4,
  2018}, pages 2979--2984.

\bibitem[{Lample et~al.(2017)Lample, Conneau, Denoyer, and
  Ranzato}]{lample2017unsupervised}
Guillaume Lample, Alexis Conneau, Ludovic Denoyer, and Marc'Aurelio Ranzato.
  2017.
\newblock Unsupervised machine translation using monolingual corpora only.
\newblock \emph{arXiv preprint arXiv:1711.00043}.

\bibitem[{Lample et~al.(2018)Lample, Conneau, Ranzato, Denoyer, and
  J{\'{e}}gou}]{DBLP:conf/iclr/LampleCRDJ18}
Guillaume Lample, Alexis Conneau, Marc'Aurelio Ranzato, Ludovic Denoyer, and
  Herv{\'{e}} J{\'{e}}gou. 2018.
\newblock \href {https://openreview.net/forum?id=H196sainb} {Word translation
  without parallel data}.
\newblock In \emph{6th International Conference on Learning Representations,
  {ICLR} 2018, Vancouver, BC, Canada, April 30 - May 3, 2018, Conference Track
  Proceedings}.

\bibitem[{Lavrenko et~al.(2002)Lavrenko, Choquette, and
  Croft}]{lavrenko2002cross}
Victor Lavrenko, Martin Choquette, and W~Bruce Croft. 2002.
\newblock Cross-lingual relevance models.
\newblock In \emph{Proceedings of the 25th annual international ACM SIGIR
  conference on Research and development in information retrieval}, pages
  175--182. ACM.

\bibitem[{Mikolov et~al.(2013{\natexlab{a}})Mikolov, Le, and
  Sutskever}]{DBLP:journals/corr/MikolovLS13}
Tomas Mikolov, Quoc~V. Le, and Ilya Sutskever. 2013{\natexlab{a}}.
\newblock \href {http://arxiv.org/abs/1309.4168} {Exploiting similarities among
  languages for machine translation}.
\newblock \emph{CoRR}, abs/1309.4168.

\bibitem[{Mikolov et~al.(2013{\natexlab{b}})Mikolov, Sutskever, Chen, Corrado,
  and Dean}]{DBLP:conf/nips/MikolovSCCD13}
Tomas Mikolov, Ilya Sutskever, Kai Chen, Greg~S Corrado, and Jeff Dean.
  2013{\natexlab{b}}.
\newblock Distributed representations of words and phrases and their
  compositionality.
\newblock In \emph{Advances in neural information processing systems}, pages
  3111--3119.

\bibitem[{Peyr{\'e} et~al.(2019)Peyr{\'e}, Cuturi
  et~al.}]{peyre2019computational}
Gabriel Peyr{\'e}, Marco Cuturi, et~al. 2019.
\newblock Computational optimal transport.
\newblock \emph{Foundations and Trends{\textregistered} in Machine Learning},
  11(5-6):355--607.

\bibitem[{Rapp(1995)}]{rapp1995identifying}
Reinhard Rapp. 1995.
\newblock Identifying word translations in non-parallel texts.
\newblock \emph{arXiv preprint cmp-lg/9505037}.

\bibitem[{Wang et~al.(2019)Wang, Zhou, Zhang, Wu, and
  Bao}]{DBLP:journals/corr/abs-1904-10294}
Zihao Wang, Datong Zhou, Yong Zhang, Hao Wu, and Chenglong Bao. 2019.
\newblock \href {http://arxiv.org/abs/1904.10294} {Wasserstein-fisher-rao
  document distance}.
\newblock \emph{CoRR}, abs/1904.10294.

\bibitem[{Xing et~al.(2015)Xing, Wang, Liu, and
  Lin}]{DBLP:conf/naacl/XingWLL15}
Chao Xing, Dong Wang, Chao Liu, and Yiye Lin. 2015.
\newblock \href {https://www.aclweb.org/anthology/N15-1104/} {Normalized word
  embedding and orthogonal transform for bilingual word translation}.
\newblock In \emph{{NAACL} {HLT} 2015, The 2015 Conference of the North
  American Chapter of the Association for Computational Linguistics: Human
  Language Technologies, Denver, Colorado, USA, May 31 - June 5, 2015}, pages
  1006--1011.

\bibitem[{Zhang et~al.(2017)Zhang, Liu, Luan, and Sun}]{zhang2017earth}
Meng Zhang, Yang Liu, Huanbo Luan, and Maosong Sun. 2017.
\newblock Earth mover’s distance minimization for unsupervised bilingual
  lexicon induction.
\newblock In \emph{Proceedings of the 2017 Conference on Empirical Methods in
  Natural Language Processing}, pages 1934--1945.

\bibitem[{Zhou et~al.(2019)Zhou, Ma, Wang, and
  Neubig}]{DBLP:conf/naacl/ZhouMWN19}
Chunting Zhou, Xuezhe Ma, Di~Wang, and Graham Neubig. 2019.
\newblock \href {https://www.aclweb.org/anthology/N19-1161/} {Density matching
  for bilingual word embedding}.
\newblock In \emph{Proceedings of the 2019 Conference of the North American
  Chapter of the Association for Computational Linguistics: Human Language
  Technologies, {NAACL-HLT} 2019, Minneapolis, MN, USA, June 2-7, 2019, Volume
  1 (Long and Short Papers)}, pages 1588--1598.

\end{thebibliography}
\bibliographystyle{acl_natbib}

\end{document}